\pdfoutput=1

\documentclass[11pt]{article}

\usepackage[]{ACL2023}

\usepackage{times}
\usepackage{latexsym}
\usepackage{graphicx}
\usepackage{enumitem}
\usepackage{float}
\usepackage{caption}
\usepackage{xcolor}
\usepackage{soul}
\usepackage{comment}
\usepackage{arydshln}
\usepackage[T1]{fontenc}

\usepackage[utf8]{inputenc}

\usepackage{microtype}

\usepackage{inconsolata}

\usepackage{subfiles}
\usepackage{booktabs, multirow}

\usepackage{xcolor}

\newcommand{\challengeName}{n2c2/UW SDOH Challenge}
\newcommand{\overlap}{any overlap}
\newcommand{\exactmatch}{exact match}
\newcommand{\mindist}{minimum distance}
\newcommand{\labelonly}{span agnostic}
\title{Prompt-based Extraction of Social Determinants of Health Using Few-shot Learning}

\newcommand{\authinfo}[2]{\fontsize{12pt}{12pt}\selectfont \textbf{#1}\textsuperscript{#2}}
\newcommand{\authinfofoot}[3]{\fontsize{12pt}{12pt}\selectfont \textbf{#1}\textsuperscript{#2}#3}
\newcommand{\affiliation}[1]{\fontsize{9pt}{9pt}\selectfont #1}

\author{
\authinfofoot{Giridhar Kaushik Ramachandran}{1}{\thanks{\quad Equal contribution}}, \authinfofoot{Yujuan Fu}{2}{\footnotemark[1]}, \authinfofoot{Bin Han}{3}{\footnotemark[1]}\\
\authinfo{Kevin Lybarger}{1}, 
\authinfo{Nicholas J Dobbins}{2},
\authinfo{\"{O}zlem Uzuner}{1},
\authinfo{Meliha Yetisgen}{2}\\
\affiliation{\textsuperscript{1} Department of Information Sciences and Technology, George Mason University, Fairfax, VA, USA} \\
\affiliation{\textsuperscript{2} Department of Biomedical Informatics \& Medical Education, University of Washington, Seattle, WA, USA}\\
\affiliation{\textsuperscript{3} Information School, University of Washington, Seattle, WA, USA}
}

\begin{document}
\maketitle

\newif\ifsubfile
\subfiletrue

\begin{abstract}

Social determinants of health (SDOH) documented in the electronic health record through unstructured text are increasingly being studied to understand how SDOH impacts patient health outcomes. In this work, we utilize the Social History Annotation Corpus (SHAC), a multi-institutional corpus of de-identified social history sections annotated for SDOH, including substance use, employment, and living status information. We explore the automatic extraction of SDOH information with SHAC in both standoff and inline annotation formats using GPT-4 in a one-shot prompting setting. We compare GPT-4 extraction performance with a high-performing supervised approach and perform thorough error analyses. Our prompt-based GPT-4 method achieved an overall 0.652 F1 on the SHAC test set, similar to the 7\textsuperscript{th} best-performing system among all teams in the n2c2 challenge with SHAC.

\end{abstract}

\section{Introduction and related work}

Social determinants of health (SDOH) are the conditions in which people work and live that impact quality of life and health \cite{cdc_2021}. Understanding SDOH can assist in clinical decision-making \cite{daniel2018addressing,friedman2018toward}. SDOH is documented in the electronic health record (EHR) through unstructured clinical narratives and structured data; however, the clinical narrative includes a more detailed description of many SDOH events. To utilize the text-encoded SDOH information in secondary use applications, including clinical decision-support systems, the SDOH information must be automatically extracted \cite{daniel2018addressing,singh2017social}. 

SDOH extraction has been explored using rule-based systems and data-driven models that use supervised learning \cite{hatef2019,patra2021,zehao2022, sifei2022} on a variety of corpora \cite{text-classification-schema1, text-classification-schema2,event-relation-schema1}. Recent SDOH extraction work utilizes large language models (LLMs) like BERT and T5, where models are fine-tuned to the SDOH extraction task \cite{lybarger2022mSpERT,msT52022}. Recent advancements in LLMs, including larger models like Generative Pretrained Transformer (GPT)-based models \cite{brown2020language, openai2023gpt4}, allow for new training paradigms, including few-shot or zero-shot learning. Recent developments in LLMs like the GPT-4 \cite{openai2023gpt4} and med-PaLM models have shown their capability to understand the clinical text and achieve/exceed human-level performance in US medical licensing exams \cite{singhal2022large}. This high performance may be attributed to (1) high model parameter counts, (2) large pre-training datasets, and (3) instruction tuning and optimization with Reinforcement Learning Human Feedback (RLHF) \cite{ouyang2022training}. Recent clinical information extraction (IE) work \cite{liu2023deid, hu2023zero} comparing BERT-based fine-tuning approaches to zero-shot learning indicates GPT models can extract entities and relations with reasonable performance; however, there are many open questions related to the use of recent LLMs, like GPT-4, in clinical IE tasks. 

In this work, we explore the extraction of SDOH using GPT-4 in a one-shot prompting setting with event-based SHAC \cite{SHAC2021}. We compare prompt-based extraction approaches with a high-performing supervised BERT-based model\cite{lybarger2022mSpERT} that has been fine-tuned to SDOH extraction from SHAC. We investigate two different one-shot prompting strategies for GPT-4, including prompts aimed at generating BRAT standoff format and inline annotations. We report an overall performance of 0.861 F1 from the fine-tuned model, evaluated on the withheld test set. The highest-performing one-shot GPT-4 approach achieved an overall 0.652 F1 for SDOH event extraction. Our initial study shows that GPT-4 can extract SDOH information from text with limited training examples.


\section{Data, Task, \& Evaluation}

\begin{figure}[hb]
    \includegraphics[width=\linewidth]{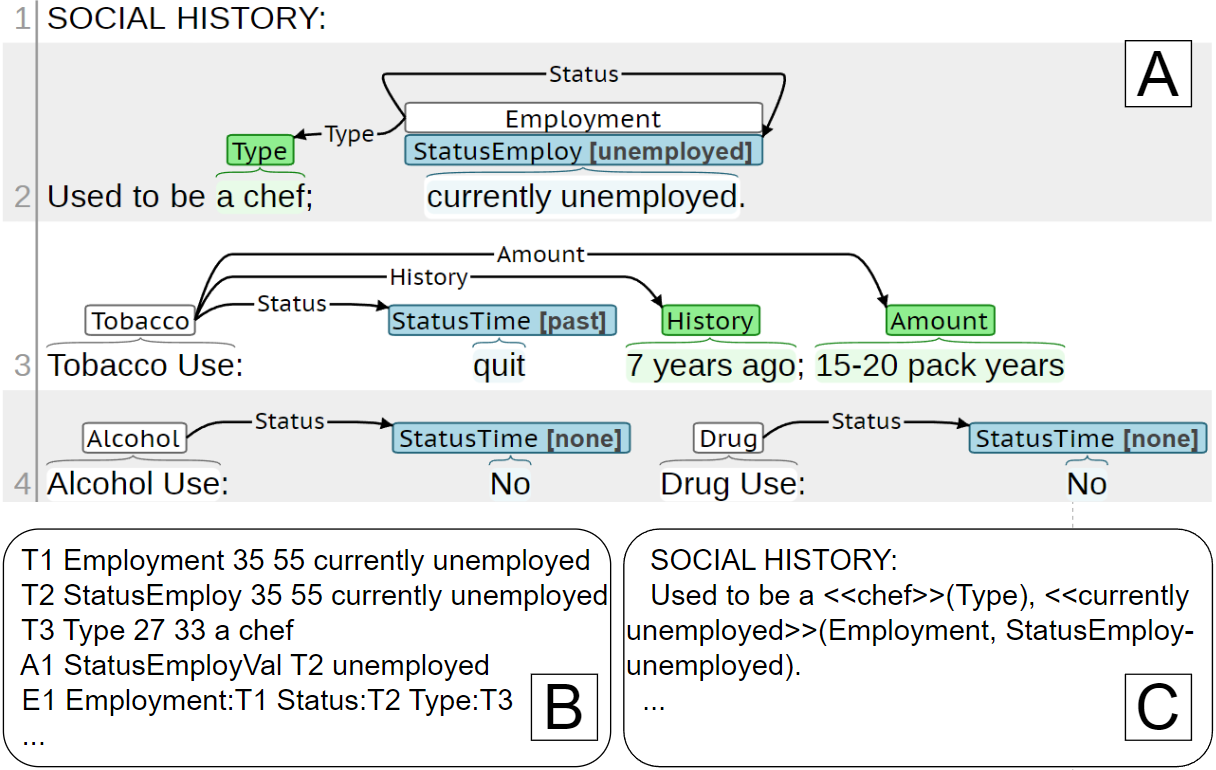}
    \caption{A. Sample note with SDOH events, visualized in the BRAT website. B. Standoff annotations in the BRAT format (.ann). C. Inline annotations. }\label{fig:brat_example}
\end{figure}

The 2022 National NLP Clinical Challenges SDOH extraction task (\challengeName{}) used SHAC for model development and evaluation \cite{n2c2}. SHAC contains 4405 de-identified social history sections of notes from MIMIC-III \cite{johnson2016mimic} and the University of Washington (UW). SHAC includes training, development, and test partitions for both sources (MIMIC-III and UW).  SHAC was annotated using BRAT \cite{stenetorp-etal-2012-brat}, a web-based annotation tool, to capture five SDOH event types: substance use (\textit{Alcohol, Drug, Tobacco}), employment status (\textit{Employment}), and living status (\textit{LivingStatus}). Figure \ref{fig:brat_example} (A) presents an annotated sample in BRAT from the SHAC UW training set. 

The \challengeName{} evaluation criteria interpret the extraction as a slot-filling task \cite{n2c2}.  Each event comprises a single trigger span and at least one required argument. \textit{Trigger} \textit{\overlap{}} equivalence requires the predicted trigger to overlap with the true trigger of the same event type. Arguments can be classified into two categories: \textit{span-only} (a multi-word span and argument type) and \textit{labeled} (a multi-word span, argument type, and \textit{subtype} label). Arguments can be equivalent only when attached to equivalent triggers. In addition to trigger \textit{\overlap{}} equivalence, \textit{span-only} argument equivalence is evaluated by \textit{\exactmatch{}}, and \textit{labeled} arguments equivalence requires the correct argument and subtype labels (\textit{\labelonly{}}) \cite{n2c2}. 

We evaluated performance using the \challengeName{} criteria, as well as on more lenient evaluation criteria that still assess the clinical meaning of extraction. In the lenient criteria, trigger equivalence is relaxed to a minimum-distance metric (\textit{\mindist{}}), where gold triggers are paired (aligned) with the closest predicted trigger of the same event type, and the closest predicted trigger is counted as a true positive. In the lenient criteria, the \textit{span-only} arguments use the \textit{\overlap{}} criteria and the \textit{labeled} arguments are evaluated as previously described.

\section{Methods}

We benchmark the SDOH extraction task using two methods: (1) multi-label variation of the Span-based Entity and Relation Transformer (SpERT)\cite{eberts2019SpERT} architecture, mSpERT \cite{lybarger2022mSpERT} benchmarked for SHAC as a high-performing fine-tuned baseline, and (2) prompt-based one-shot learning with GPT-4. Inspired by performance gains of few-shot learning, relative to zero-shot learning in prior work \cite{brown2020language, lievin2022can}, we use one-shot prompting with GPT-4 for the SDOH extraction task in this short study. We experiment with two distinct output formats - (1) BRAT-style standoff annotations (GPT-standoff) and (2) Inline annotations (GPT-inline).

We conduct the GPT-4 one-shot experiments through OpenAI's GPT-4 Chat Completion Application Programming Interface (API)\footnote[1]{\url{https://platform.openai.com/docs/api-reference/chat}}, because of GPT-4's proprietary nature and significant hardware requirements. The API allows users to provide instructions via three role variables. Our prompts are structured in the following order: 
\begin{enumerate}[leftmargin=0.45cm]
  \setlength{\itemsep}{0pt}
  \setlength{\parskip}{0pt}
  \setlength{\parsep}{0pt}
    \item \textbf{\textit{system}}: defines the desired role, personality traits, and task instructions for GPT-4. We use the \textit{system} variable to assign GPT-4 the role of an annotator along with the paraphrased annotation guideline. 
    \item \textbf{\textit{user}}: provides an example note for one-shot learning.
    \item \textbf{\textit{assistant}}: provides the gold annotations for the example note in \textit{user}. This is an example prediction for one-shot learning.

\end{enumerate}
Following the above definitions, we end with a \textit{user} message containing a note to be annotated and indicate the \textit{assistant} should respond.  We randomly sample a note from a subset of the SHAC-UW training set containing all five SDOH event types. Appendix \ref{prompting_method} contains formats and snippets of our prompts. 

\subsection{GPT-standoff}
To assess GPT-4's ability to comprehend the task and generate structured outputs, we prompt the model to generate predictions in the BRAT  standoff format \cite{stenetorp-etal-2012-brat} used by SHAC. The BRAT  standoff format includes pairs of text (*.txt) and annotation (*.ann) files. The event annotation is characterized by three BRAT annotation frames in the annotation file: (1) \textit{Text bounds ($T$)} include a text span (e.g. ``currently unemployed''), span label (e.g. `Employment'), and character indices (e.g. ``35 55'') for marking both triggers and arguments; (2) \textit{Attributes ($A$)} adds a subtype label to $T$, and (3) \textit{Events ($E$)} characterize an SDOH event through linking a trigger and at least one argument. A visual representation is provided in Figure \ref{fig:brat_example} (B).

We provide the paraphrased annotation guideline in the \textit{system} and the note via \textit{user} variables and elicit GPT-4 to output the desired annotation file through the \textit{assistant} variable. Our preliminary experimentation indicated that though GPT-4 was able to correctly extract relevant text spans, it had some shortcomings: (1) some generated lines did not conform to the BRAT standoff format, and (2) the generated character indices did not correspond with the identified text spans. We post-processed the generated outputs to ensure compliance with the BRAT standoff format and updated the character indices to correspond with the first occurrence of the generated text span (<3\% spans occur more than once). Please refer to Appendix \ref{Post-processing} for original and post-processed examples.

\subsection{GPT-inline}
Prior fine-tuned IE work utilized inline markers to infuse entity information in the body of narratives \cite{msT52022, phan2021scifive}. In our work, we instruct GPT-4 to generate a version of the note with inline markers that identify the SDOH triggers and arguments. These markers encode all spans inside double-angle brackets (``<< >>''), with trigger, argument, and subtype labels appended (Figure \ref{fig:brat_example}, (C)). Similar to the GPT-standoff model, the GPT-inline model elicits GPT-4 the desired annotation format through the \textit{assistant} variable. This method does not prompt the model to make trigger-argument connections, and we use a heuristic search to associate each argument with the nearest event trigger, constrained by the allowable trigger-argument connections defined by the annotation guideline (See details in the Appendix. \ref{marker-connection}). The GPT-inline output is post-processed into BRAT standoff format for evaluation.





\section{Results and Discussion}

Table \ref{tab:comparison_with_n2c2} contains the overall performance of the prompt-based GPT-4 models on the withheld SHAC-UW test set, which was the evaluation data for Subtask C of \challengeName{} \cite{n2c2}.  The GPT-standoff and GPT-inline models achieve an overall F1 of 0.625 and 0.652, respectively. This performance is much lower than the mSpERT model and the highest-performing n2c2 systems, which utilized the entire training set in supervised model fine-tuning. GPT-inline achieved performance similar to the 7\textsuperscript{th} best n2c2 system from IBM, which utilized BERT \cite{n2c2}. This one-shot performance indicates that the natural language understanding capabilities of GPT-4 allow the prompt-based methods to leverage the annotation schema and achieve moderate performance. The results suggest that fine-tuning at least a portion of the training set may be needed to achieve high performance. We also observed that a generative architecture could lead to a new set of errors, and some of them may be eliminated in post-processing.

\begin{table}[ht]
\centering
\small

\begin{tabular}{lccc}
\hline
\textbf{Method} & \textbf{P} & \textbf{R} & \textbf{F1} \\ \hline
\multicolumn{4}{l}{\textbf{Fine-tuned}} \\
Microsoft (T5) & 0.891 & 0.887 & 0.889 \\
CHOP (BERT) & 0.874 & 0.888 & 0.881 \\
mSpERT & 0.868 & 0.854 & 0.861 \\
... & ... & ... & ... \\
IBM (BERT) & 0.538 & 0.788 & 0.640 \\
\multicolumn{4}{l}{\textbf{GPT-4 one-shot + post-processing}} \\

GPT-standoff & 0.621 & 0.628 & 0.625 \\
GPT-inline & 0.650 & 0.654 & 0.652 \\ \hline
\end{tabular}
\caption{Comparison of overall micro-averaged performance of SDOH triggers and arguments between select top-performing models in the n2c2 challenge, mSpERT (fine-tuned baseline), and the GPT-4 models.}
\label{tab:comparison_with_n2c2}
\end{table}

\begin{table*}[ht]
\centering
\small

\begin{tabular}{llccccccc}%
\hline
\multirow{2}{*}{\textbf{Field}} & \multirow{2}{*}{\textbf{Argument}} & \multicolumn{1}{l}{\multirow{2}{*}{\textbf{\begin{tabular}[c]{@{}l@{}}\# True \\ Labels\end{tabular}}}} & \multicolumn{3}{c}{\textbf{n2c2 Evaluation (F1)}} & \multicolumn{3}{c}{\textbf{Lenient Evaluation (F1)} } \\ \cline{4-9} 
 & \multicolumn{2}{l}{} & \textbf{mSpERT} & \begin{tabular}[c]{@{}l@{}}\textbf{GPT-}\\\textbf{standoff}\end{tabular} & \begin{tabular}[c]{@{}l@{}}\textbf{GPT-}\\\textbf{inline}\end{tabular} & \textbf{mSpERT} & \begin{tabular}[c]{@{}l@{}}\textbf{GPT-}\\\textbf{standoff}\end{tabular} & \begin{tabular}[c]{@{}l@{}}\textbf{GPT-}\\\textbf{inline}\end{tabular}\\ \hline
\multicolumn{2}{l}{\textbf{Trigger}} & \multicolumn{1}{l|}{} & \multicolumn{1}{l:}{} & \multicolumn{1}{l}{} & \multicolumn{1}{l|}{} & \multicolumn{1}{l:}{} & \multicolumn{1}{l}{} & \multicolumn{1}{l}{} \\
Alcohol & - & \multicolumn{1}{c|}{403} & \multicolumn{1}{c:}{0.964} & 0.861 & \multicolumn{1}{c|}{0.938*} & \multicolumn{1}{c:}{0.967} & 0.972* & 0.952 \\
Drug & - & \multicolumn{1}{c|}{473} & \multicolumn{1}{c:}{0.929} & 0.824 & \multicolumn{1}{c|}{0.861*} & \multicolumn{1}{c:}{0.942} & 0.935* & 0.898 \\
Tobacco & - & \multicolumn{1}{c|}{434} & \multicolumn{1}{c:}{0.963} & 0.825 & \multicolumn{1}{c|}{0.917*} & \multicolumn{1}{c:}{0.970} & 0.965* & 0.939 \\
Employment & - & \multicolumn{1}{c|}{153} & \multicolumn{1}{c:}{0.908} & 0.803* & \multicolumn{1}{c|}{0.709} & \multicolumn{1}{c:}{0.915} & 0.921* & 0.766 \\
LivingStatus & - & \multicolumn{1}{c}{354} & \multicolumn{1}{c}{0.886} & 0.590 & \multicolumn{1}{c}{0.749*} & \multicolumn{1}{c}{0.903} & 0.844* & 0.811 \\
\multicolumn{2}{l}{\textbf{Labeled Argument}} & \multicolumn{1}{l|}{} &  & \multicolumn{1}{:l}{} & \textbf{} & \multicolumn{1}{|l}{} & \multicolumn{1}{:l}{} & \multicolumn{1}{l}{} \\
Alcohol & StatusTime & \multicolumn{1}{c|}{403} & \multicolumn{1}{c:}{0.913} & 0.763 & \multicolumn{1}{c|}{0.734} & \multicolumn{1}{c:}{0.913} & 0.856* & 0.750 \\
Drug & StatusTime & \multicolumn{1}{c|}{473} & \multicolumn{1}{c:}{0.857} & 0.706* & \multicolumn{1}{c|}{0.646} & \multicolumn{1}{c:}{0.868} & 0.783* & 0.673 \\
Tobacco & StatusTime & \multicolumn{1}{c|}{434} & \multicolumn{1}{c:}{0.917} & 0.694 & \multicolumn{1}{c|}{0.738*} & \multicolumn{1}{c:}{0.926} & 0.813* & 0.764 \\
Employment & StatusEmploy & \multicolumn{1}{c|}{153} & \multicolumn{1}{c:}{0.868} & 0.657 & \multicolumn{1}{c|}{0.627} & \multicolumn{1}{c:}{0.875} & 0.759* & 0.679 \\
\multirow{2}{*}{LivingStatus} & StatusTime & \multicolumn{1}{c|}{354} & \multicolumn{1}{c:}{0.833} & 0.572 & \multicolumn{1}{c|}{0.709*} & \multicolumn{1}{c:}{0.850} & 0.787* & 0.760 \\
 & TypeLiving & \multicolumn{1}{c}{354} & \multicolumn{1}{c}{0.871} & 0.560 & \multicolumn{1}{c}{0.725*} & \multicolumn{1}{c}{0.891} & 0.770* & 0.774 \\
\multicolumn{2}{l}{\textbf{Span-only Argument}} & \multicolumn{1}{l|}{} &  & \multicolumn{1}{:l}{} &  & \multicolumn{1}{|l}{} & \multicolumn{1}{:l}{} & \multicolumn{1}{l}{} \\
Alcohol & All types & \multicolumn{1}{c|}{178} & \multicolumn{1}{c:}{0.699} & 0.388* & \multicolumn{1}{c|}{0.172} & \multicolumn{1}{c:}{0.783} & 0.694* & 0.354 \\

Drug & All types & \multicolumn{1}{c|}{418} & \multicolumn{1}{c:}{0.625} & 0.219* & \multicolumn{1}{c|}{0.104} & \multicolumn{1}{c:}{0.688} & 0.426 & 0.381 \\
Tobacco & All types & \multicolumn{1}{c|}{375} & \multicolumn{1}{c:}{0.775} & 0.420* & \multicolumn{1}{c|}{0.322} & \multicolumn{1}{c:}{0.830} & 0.714* & 0.537 \\
Employment & Duration, History, Type & \multicolumn{1}{c|}{96} & \multicolumn{1}{c:}{0.675} & 0.169 & \multicolumn{1}{c|}{0.109} & \multicolumn{1}{c:}{0.735} & 0.677* & 0.500 \\
LivingStatus & Duration, History & \multicolumn{1}{c|}{11} & \multicolumn{1}{c:}{0.421} & 0.063 & \multicolumn{1}{c|}{0.074} & \multicolumn{1}{c:}{0.526} & 0.159 & 0.105 \\ \midrule
\multicolumn{2}{c}{\textbf{Overall}} & \multicolumn{1}{c|}{5066} & \multicolumn{1}{c:}{0.861} & 0.625 & \multicolumn{1}{c|}{0.652*} & \multicolumn{1}{c:}{0.882} & 0.791* & 0.728 \\ \hline
\end{tabular}
\caption{Micro-averaged F1 comparison between mSpERT and GPT-4 one-shot models. SpERT outperforms the GPT-4 one-shot models in all trigger and argument extraction (with the exception of \textit{Alcohol} and \textit{Employment} triggers). For better readability, we only mark performance significance among GPT-4 one-shot models. * indicates performance significance among GPT-4 one-shot methods, with 10,000 bootstrap samples and a p-value threshold of 0.05. }
\label{tab:detailed_results}
\end{table*}

Table \ref{tab:detailed_results} contains trigger and argument micro-averaged F1 scores for each event type using the n2c2 and lenient evaluation criteria. Comparing overall performances (last row in the table), mSpERT outperforms both our GPT-4 models. But the performance gap between the fine-tuned model and the one-shot GPT models is smaller from the n2c2 evaluation to the lenient evaluation, which can largely be attributed to higher trigger extraction performance, as argument equivalence requires trigger equivalence in both evaluations. The GPT-standoff model can identify the presence of SDOH events, but the identified triggers may not overlap with the gold trigger. The lenient evaluation only requires the same trigger type present in the social history text. The relatively lower performance for the GPT-inline model in \textit{Employment} trigger extraction can be attributed to the model frequently identifying an \textit{StatusEmploy} labeled argument without predicting an \textit{Employment} trigger. The GPT-inline extractions may capture meaningful employment information but do not adhere to the annotation guidelines. For both GPT models, the extraction of \textit{LivingStatus} triggers is relatively more challenging. Although the notes contain many plausible candidate spans for \textit{LivingStatus} triggers, these spans were not annotated in SHAC since they did not contain information to resolve the associated \textit{TypeLiving} labeled argument. The GPT models capture these false positive \textit{LivingStatus} triggers often without \textit{TypeLiving} labeled arguments. For argument extraction, GPT-standoff significantly outperforms GPT-inline in four arguments under the n2c2 evaluation and eight arguments in lenient evaluation. The GPT-inline does not link annotated arguments to triggers, and a distance metric (character count) is used to link them, which contributes to the GPT-in-line's relatively lower performance. The labeled arguments are required for each event, and we observe that labeled argument performance is ~0.1 F1 lower than the corresponding trigger performance. When multiple substance events are present in a note with differing \textit{StatusTime} labels (e.g. current and past), we observe the GPT-standoff model tends to output the same \textit{StatusTime} label for all substance events. The GPT-inline model correctly captures both substance trigger and \textit{StatusTime} spans with the correct label but fails to correctly link the triggers with the right \textit{StatusTime} spans because multiple \textit{StatusTime} spans can have the same distance to a trigger span.

\section{Conclusions}

We investigate the efficacy of two prompt-based approaches for extracting SDOH from social history sections using GPT-4. Although the supervised model achieves higher performance, our findings indicate that GPT-4's one-shot learning capabilities serve as a promising starting point for extracting SDOH events without the need for annotated data. Possible gains in future work may be achieved with a combination of few-shot and active learning.

\section{Limitations}

We only explored one-shot prompting strategies with GPT-4. More examples (few-shot) may improve performance. We prompted GPT-4 with only a single randomly selected sample that included all of the annotated event types. Our post-processing included simple rules to process the generated output and may be improved. The quality of the sample and the selection method may influence performance. We explored two prompting styles. Future work could explore more prompting methods such as question \& answering and chain-of-thought \cite{wei2023chainofthought} and fine-tuning non-proprietary LLMs.

\section{Ethics statement}

Our experimentation utilized OpenAI API to extract SDOH information from SHAC with GPT-4. SHAC is a fully de-identified corpus of social history sections. The use of such external API/models could introduce ethical problems related to privacy, identifiability, and other unintended consequences if the data sets are not fully de-identified. Additionally, a careful examination is needed to assess potential bias in LLMs for extracting SDOH prior to implementing real-life secondary use applications. 
We received approval from the Institutional Review Board (IRB) prior to conducting the presented research. As our GPT-4 one-shot experiments are conducted on the SHAC-UW test set, broader use of the model may need necessary precautions.

\section{Acknowledgements}
This work was supported in part by the National Institutes of Health and the National Library of Medicine (NLM) (Grant numbers R01 CA248422-01A1, R15 LM013209). The content is solely the responsibility of the authors and does not necessarily represent the official views of the National Institutes of Health. 

\bibliography{custom}
\bibliographystyle{acl_natbib}

\appendix
\onecolumn
\section{Appendix}

\subsection{The SHAC annotation schema} \label{annotation_schema}

\begin{table}[H]
\centering
\begin{tabular}{llll}
\hline
\textbf{Event type} & \textbf{Argument type} & \textbf{Argument subtypes} & \textbf{Span examples} \\ \hline
\multirow{6}{*}{\begin{tabular}[c]{@{}l@{}}Alcohol, Drug, \\ \& Tobacco\end{tabular}} & StatusTime* & \{none, current, past\} & “denies,” “no history” \\ 
 & Amount & – & “1 pack”, “3 drinks” \\  
& Duration & – & “since last week” \\ 
 & Frequency & – & “1-2x/week”, “daily” \\ 
 & History & – & “when he was young” \\ 
  & Type & – & “smokeless,”\\ &&&“methamphetamine” \\ 
 \hline
\multirow{4}{*}{Employment} & StatusEmploy* & \begin{tabular}[c]{@{}l@{}}\{employed, unemployed, retired, \\ on disability, student, homemaker\}\end{tabular} & “works,” “unemployed” \\ 
 & Duration & – & “since last week” \\ \ 
 & History & – & “10 years ago” \\ 
 & Type & – & “remote office work” \\ \hline
\multirow{4}{*}{LivingStatus} & StatusTime* & \{current, past, future\} & “lives,” “lived” \\ 
 & TypeLiving* & \begin{tabular}[c]{@{}l@{}}\{alone, with family, with others, \\ homeless\}\end{tabular} & “with husband,” “alone” \\ 
 & Duration & – & “since he was 12” \\ 
 & History & – & “until 2 years ago” \\ \hline
\end{tabular}
\caption{Annotation guideline summary. *indicates the argument is required}
\label{tab: annotation_schema}
\end{table}

\subsection{Prompt Methods} \label{prompting_method}

The exact prompting messages are listed below. The annotation guideline is used for the SHAC dataset creation. We removed all the annotation examples in the guideline, as some of the examples are from MIMIC-III. We also remove invalid references to tables. 

\subsubsection{Message 1 - \textit{System}}
\paragraph{GPT-4 role definition for standoff annotation}
\hspace{1em}\newline
You are an expert medical annotator and understand the BRAT standoff format very well.  You are given a document that contains the following list of entities and events:''
\paragraph{GPT-4 role definition for inline annotation}
\hspace{1em}\newline
``You are an expert medical annotator who adds annotations as inline markers in documents. You are given a document to annotate the following list of entities, events, and attributes:
\paragraph{Annotation guideline}
\hspace{1em}\newline
The annotation involves the identification of SDOH events, where each SDOH event is represented by a trigger and set of entities. The trigger consists of a multi-word span (word or phrase) and a label indicating the type of SDOH (e.g. employment or tobacco use). All annotated phenomena are defined in terms of the span (words associated with phenomena) and the span type (e.g. amount, status, etc.), and some annotated phenomena, like status, will also include a span label (e.g. current or past). If an annotated span includes a noun, the selected span should include the entire applicable noun phrase. If an annotated span includes a verb, the selected span should include applicable adverbs. The inclusion of entire noun phrases and relevant adverbs is applicable to all annotated phenomena.
You are given a document that contains the following list of  events and entities:
\newline
\newline
[EVENTS]
\begin{itemize}
\item 
{Alcohol}: 
{Alcohol} event indicates the usage of alcohol. The trigger span should be a noun phrase describing a general substance type, like ``alcohol” if present. Alcohol may be abbreviated ETOH in chart notes. If general substance types are not present, the span should be a relevant verb (e.g. "drink") or a noun phrase describing a specific substance type (e.g. "beer"). Examples of {Alcohol} spans include: ["alcohol", "beer", "drink"].
...

\end{itemize}

[ENTITIES]

<Span only entities>
\begin{itemize}
\item 
{Amount}: 
Linked to substance use events ({Alcohol}, {Drug}, {Tobacco}). {Amount} indicates the quantity of the substance used by the patient  (e.g. “three drinks” or “2 packs”) and do not include qualitative descriptions (e.g. "significant" or “heavy”) . Amount spans should not include frequency information, and frequency and amount spans should not overlap. Examples of {Amount} entities include: ["one glass", "up to three drinks", "1 ppd", "half a pack", "4 packs"].

...

\end{itemize}

\subsubsection{Message 2 - \textit{User}}
\paragraph{Annotation instructions for standoff annotation}
\hspace*{1em}\newline
Based on this annotation guideline, please annotate the following document in the BRAT standoff format.

\paragraph{Annotation instructions for inline annotation}
\hspace*{1em}\newline
Based on this annotation guideline, please annotate the following document with inline markers. 
\paragraph{An example document}
\hspace*{1em}\newline
HABITS: \newline
Tobacco Use: denies   Alcohol Use: denies   Drug Use: denies \newline
\newline
SOCIAL HISTORY: \newline
Residence: [LOCATION] with husband and kids \newline
Job: no longer works \newline
 
\subsubsection{Message 3 - \textit{Assistant}}
\paragraph{Annotated example in standoff annotation}
\hspace*{1em}\newline
T1	LivingStatus 88 97	Residence \newline
T2	TypeLiving 110 131	with husband and kids \newline
T3	StatusTime 88 97	Residence \newline
... \newline

T11	StatusEmploy 137 152	no longer works \newline
E1	LivingStatus:T1 Type:T2 Status:T3 \newline
E2	Tobacco:T4 Status:T6 \newline
... \newline

A1	TypeLivingVal T2 with\_family \newline
A2	StatusTimeVal T3 current \newline
... 

\paragraph{Annotated example for inline annotation}
\hspace*{1em}\newline
HABITS: \newline
<<Tobacco Use>>(Tobacco): <<denies>>(StatusTime-none)  <<Alcohol Use>>(Alcoho): <<denies>>(StatusTime-none)   <<Drug Use>>(Drug): <<denies>>(StatusTime-none)\newline
\newline
SOCIAL HISTORY:\newline
<<Residence>>(LivingStatus, StatusTime-current): [LOCATION] <<with husband and kids>>(TypeLiving-with\_family) \newline
<<Job>>(Employment): <<no longer works>>(StatusEmploy-unemployed)
\subsubsection{Message 4 - \textit{User}}
The annotation instructions for standoff and inline annotations are the same as in Message 2.

\subsection{Examples of Post-processing} \label{Post-processing}
The example document before and after post-processing can be found in Table. \ref{tab:w/wo-post}
\begin{table}[h]
\centering
\small
\begin{tabular}{lll}
\hline
\textbf{} &
  \textbf{Before Post-processing} &
  \textbf{After Post-processing} \\ \hline
\textbf{\begin{tabular}[c]{@{}l@{}}GPT-\\ standoff\end{tabular}} &
  \begin{tabular}[c]{@{}l@{}}T1	LivingStatus \textcolor{red}{26 35}	Residence\\ T2	TypeLiving \textcolor{red}{43 64}	with husband and kids \\ T3	StatusTime \textcolor{red}{75 97}	Residence \\ ...\\ T11	StatusEmploy \textcolor{red}{110 123}	no longer works\\ A1	TypeLivingVal T2 with\_family \\  A2	StatusTimeVal T3 current\\...\\ E1	LivingStatus:T1 Type:\textcolor{red}{T14} Status:T3 \\ ...\end{tabular} &
  \begin{tabular}[c]{@{}l@{}}T1	LivingStatus \textcolor{green}{88 97}	Residence\\ T2	TypeLiving \textcolor{green}{110 131}	with husband and kids\\ T3	StatusTime \textcolor{green}{88 97}	Residence\\ ...\\ T11	StatusEmploy \textcolor{green}{137 152}	no longer works \\ A1	TypeLivingVal T2 with\_family \\  A2	StatusTimeVal T3 current\\...\\ E1	LivingStatus:T1 Status:T3 \\ ...\end{tabular} \\ \cline{1-1}

\textbf{\begin{tabular}[c]{@{}l@{}}GPT-\\ inline\end{tabular}} &
  \begin{tabular}[c]{@{}l@{}}...\\ SOCIAL HISTORY:\\ <<Residence>>(LivingStatus, StatusTime-\\ current): {[}LOCATION{]} <<with husband and\\  kids>>(TypeLiving-with\_family)\\ <<Job>>(Employment) \textcolor{red}{Status}: <<no longer \\works>\textgreater (StatusEmploy-unemployed).\end{tabular} &
  \begin{tabular}[c]{@{}l@{}}...\\ SOCIAL HISTORY:\\ <<Residence>>(LivingStatus, StatusTime-\\ current): {[}LOCATION{]} <<with husband and \\kids >>(TypeLiving-with\_family)\\ <<Job>>(Employment): <<no longer works\\ >\textgreater(StatusEmploy-unemployed)\end{tabular} \\ \hline
\end{tabular}
\caption{GPT-4 responses before and after post-processing.}
\label{tab:w/wo-post}
\end{table}
\vspace{-0.4cm}

\subsection{Heuristics for Trigger-argument Connections in the GPT-inline Outputs}\label{marker-connection}
We use the sample example in Table. \ref{tab:w/wo-post} to demonstrate our heuristics for finding the trigger-arguments from GPT-inline outputs and output to BRAT standoff format. For example, after post-processing, the $T$ and $A$ arguments can be directly extracted:

\begin{tabbing}
\hspace{30pt}\=\hspace{30pt}\=\kill
\> T1	LivingStatus 88 97	Residence \\
\> T2	TypeLiving 110 131	with husband and kids \\
\> A1	TypeLivingVal T2 with\_family \\
\> T3	StatusTime 88 97	Residence \\
\> A2	StatusTimeVal T3 current \\
\> T4	Employment 132 135	Job \\
\> T5	StatusEmploy 137 152	no longer works \\
\> A6	StatusEmployVal T5 unemployed \\
\> ...
\end{tabbing}
The above examples contain two trigger spans: T1 and T4. For the rest argument spans, we want to link each of them to its closest trigger span, constrained by the annotation guideline. For example, the distance between the argument T2 and the trigger T1 is 132 (T4 start) - 131 (T2 end) = 1 (character index), and the distance between T2 and the trigger T4 is 131 (T2 start) - 97 (T1 end) = 34 (character index). T1 is closer to the trigger T4. However, because the \textit{TypeLiving} argument can only be attached to the \textit{LivingStatus} trigger, T1 is attached to its closet  \textit{LivingStatus} trigger T4. Note that it is possible that a trigger does not contain any arguments or an argument is not attached to any trigger in GPT-inline outputs. Arguments in the above example can be summarized into BRAT events as:
\begin{tabbing}
\hspace{30pt}\=\hspace{30pt}\=\kill
\> E1	LivingStatus:T1 Type:T2 Status:T3 \\
\> E2	Employment:T4 Status:T5 \\
\> ...
\end{tabbing}

\end{document}